\documentclass[10pt,twocolumn,letterpaper]{article}

\usepackage{iccv}
\usepackage{times}
\usepackage{epsfig}
\usepackage{graphicx}
\usepackage{amsmath}
\usepackage{amssymb}

\usepackage{booktabs}
\usepackage{multirow}
\usepackage{pifont}
\usepackage{makecell}

\usepackage[accsupp]{axessibility}  % Improves PDF readability for those with disabilities.

\usepackage{marvosym}
\usepackage[pagebackref=true,breaklinks=true,letterpaper=true,colorlinks,bookmarks=false]{hyperref}

\usepackage[capitalize]{cleveref}
\crefname{section}{Sec.}{Secs.}
\Crefname{section}{Section}{Sections}
\Crefname{table}{Table}{Tables}
\crefname{table}{Tab.}{Tabs.}

\iccvfinalcopy % *** Uncomment this line for the final submission

\def\iccvPaperID{6790} % *** Enter the ICCV Paper ID here

\ificcvfinal\fi

\begin{document}

%%%%%%%%% TITLE
\title{3D Semantic Subspace Traverser: Empowering 3D Generative Model \\
with Shape Editing Capability}

\author{
Ruowei Wang \qquad Yu Liu \qquad Pei Su \qquad Jianwei Zhang \qquad Qijun Zhao\textsuperscript{\Letter} \\
Sichuan University\\
% Institution1 address\\
{\tt\small \{ruoweiwang, liuyuvincent, supei\}@stu.scu.edu.cn   \qquad \{zhangjianwei, qjzhao\}@scu.edu.cn}
% For a paper whose authors are all at the same institution,
% omit the following lines up until the closing ``}''.
% Additional authors and addresses can be added with ``\and'',
% just like the second author.
% To save space, use either the email address or home page, not both
}

\maketitle
% Remove page # from the first page of camera-ready.
\ificcvfinal\thispagestyle{empty}\fi

%%%%%%%%% ABSTRACT
\begin{abstract}
% 编辑的范式改变了后，这里说明方式也应该需要改变。
Shape generation is the practice of producing 3D shapes as various representations for 3D content creation. Previous studies on 3D shape generation have focused on shape quality and structure, without or less considering the importance of semantic information. Consequently, such generative models often fail to preserve the semantic consistency of shape structure or enable manipulation of the semantic attributes of shapes during generation. In this paper, we proposed a novel semantic generative model named 3D Semantic Subspace Traverser that utilizes semantic attributes for category-specific 3D shape generation and editing. Our method utilizes implicit functions as the 3D shape representation and combines a novel latent-space GAN with a linear subspace model to discover semantic dimensions in the local latent space of 3D shapes. Each dimension of the subspace corresponds to a particular semantic attribute, and we can edit the attributes of generated shapes by traversing the coefficients of those dimensions. Experimental results demonstrate that our method can produce plausible shapes with complex structures and enable the editing of semantic attributes. The code and trained models are available at \href{https://github.com/TrepangCat/3D_Semantic_Subspace_Traverser}{https://github.com/TrepangCat/3D\_Semantic\_Subspace\_Tra\\verser}
\end{abstract}

%%%%%%%%% BODY TEXT
\section{Introduction}

Building a generative model to synthesize 3D shapes is a predominant topic in computer graphics and 3D vision since it is widely used in applications such as 3D content creation/reconstruction \cite{mmvad, sketch2mesh, fully-understanding-generic-objects-modeling-segmentation-and-reconstruction,2d-gans-meet-unsupervised-single-view-3D-reconstruction}, autonomous driving \cite{autodrive1, autodrive2, autodrive3}, AR/VR \cite{ar_vr}, and robotics \cite{robotics1, robotics2}. Typically, a 3D shape generative model takes a random vector as input and produces a 3D shape as output. To evaluate the effectiveness of generative models, researchers consider three aspects: quality, diversity, and controllability. The ideal 3D shape generative model should produce visually realistic and diverse 3D shapes while also enabling local structure manipulation to achieve controllability.

\begin{figure}[t]
	\centering
% 	\fbox{\rule{0pt}{2in} \rule{0.9\linewidth}{0pt}}
	\includegraphics[width=\linewidth]{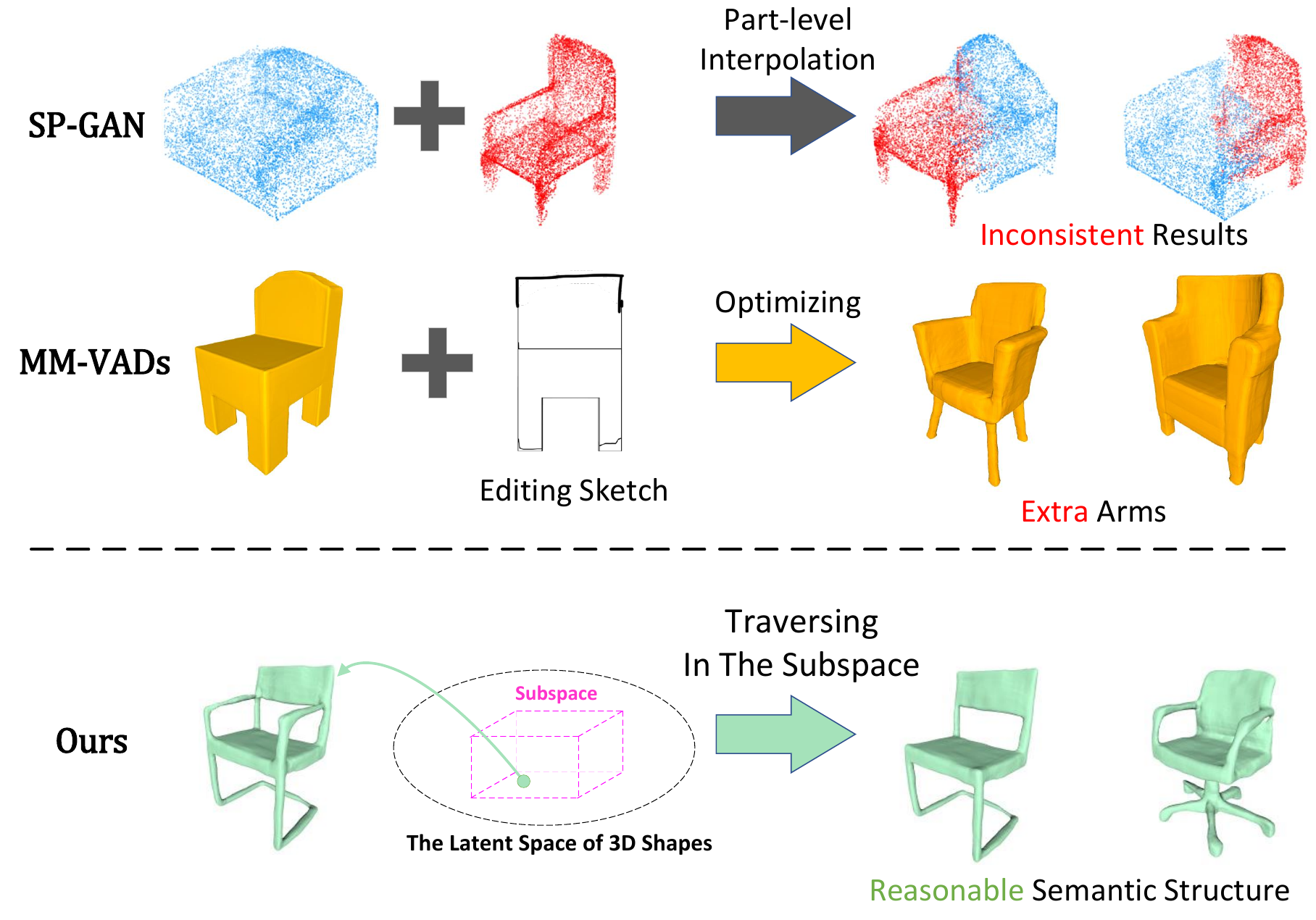}
	\caption{Different Editing Paradigm. Since SP-GAN is a point cloud generative model, we present its results as point clouds. As illustrated, the first row shows a typical method using part-level interpolation for shape editing. The second row is an example of the optimization-based method for shape editing. The last row shows how we traverse the local linear subspace to produce edited shapes with reasonable semantic structures.
	}
	\label{fig:consistency}
\end{figure}

Previous 3D shape generative models like variational autoencoder (VAE) \cite{atlas-net, GRASS, editvae, dsg-net, pq-net}, generative adversarial network (GAN) \cite{3D-gan, pc-gan, spgan, im-gan, grid-im-gan, shapegan, sdfstylegan}, flow-based models \cite{pointflow, dpf, softflow}, and diffusion models \cite{luo2021diffusion, lion, luo2021diffusion, pvd} have been applied to various 3D representations, including point clouds \cite{pc-gan}, voxels \cite{3D-gan}, mesh \cite{atlas-net} and so on. By adopting the rapidly developed implicit function-based shape representation \cite{im-gan, deepls, deepsdf}, the generated shapes show promising quality and diversity \cite{sdfstylegan}. However, the ability to control and manipulate the local structures of generated shapes during the generation process remains a challenging task.
It is crucial to develop methods for effective shape control that allows for the modification of local structures during the generation process.

Previous methods that edit local shapes can be classified into two groups: interpolation-based methods \cite{g2l-gan, pq-net, spgan, StructureNet, SPAGHETTI, editvae, learning-implicit-functions-for-dense-3D-shape-correspondence-of-generic-objects, learning-implicit-functions-for-topology-varying-dense-3D-shape-correspondence} and optimization-based methods \cite{dualsdf, dif-net, shapecrafter, implicit_text_3D, mmvad, sketch2mesh}. The interpolation-based ones utilize part-level interpolation to edit shapes. However, since it does not consider the semantic relation between shape parts, it produces inconsistent results (see in the first row of \cref{fig:consistency}). The optimization-based methods take the editing inputs, which are sketches, point moving operations, and so on, to optimize shapes. But its results focus on matching the editing inputs while ignoring other shape parts (see in the second row of \cref{fig:consistency}). In light of this, we propose a new paradigm that utilizes the learned semantic information in the subspace (see in the last row of \cref{fig:consistency}). During the generation process, we can edit shapes by traversing semantic dimensions in the subspace to produce diverse results. More details are illustrated as follows:

We present 3D Semantic Subspace Traverser, a novel semantic generative model for 3D shape generation and editing by leveraging the proposed local linear subspace models and the latent-space GAN. We adopt deep implicit function as the 3D representation and apply our GAN to the latent shape space, which is produced by a VAE. The GAN generates shape features in the form of a shape code grid, which can be decoded into 3D shapes by the VAE decoder. To empower our 3D generative model with shape editing capability, we embed the local linear subspace models into the GAN. Those local linear subspace models can discover semantic dimensions from feature maps in an unsupervised manner. By traversing along those semantic dimensions, our GAN can control the semantic attributes of the generated shapes, as shown in \cref{fig:consistency}. The proposed 3D Semantic Subspace Traverser generates plausible 3D shapes and enables semantically editing of 3D shapes, as evidenced by quantitative and qualitative results.

To summarize, the contributions of this work are:
\begin{itemize}
\item We propose 3D Semantic Subspace Traverser, a novel semantic generative model that facilitates semantic shape editing for 3D shapes.
\item We present a new latent-space GAN that leverages shape code grids to enable implicit-function-based 3D generation, producing 3D shapes with diverse topological structures.
\item We introduce a 3D local linear subspace model to unsupervisedly and progressively mine interpretable and controllable dimensions from generator layers.
\item Superior performance has been achieved in both 3D shape generation and editing.
\end{itemize}

\section{Related Work}

\subsection{Generative Models for 3D Shape}

\noindent\textbf{GAN-based methods.} 
3D-GAN \cite{3D-gan} first introduces the GAN network to the voxel-based 3D generation, since the format of the voxel is suitable for the neural network. G2L-GAN \cite{g2l-gan} produces the voxels with the part labels. PAGENet \cite{pagenet} generates the voxel parts of the shapes and assembles them together. PC-GAN \cite{pc-gan} first uses the GAN network for the point cloud and proposes the latent-space GAN (l-GAN) which trains the GAN on the latent space of a pre-trained autoencoder. Tree-GAN \cite{tree-gan} uses tree structures \cite{treestructure1} for the GAN network with graph convolutions. Progressive-PCGAN \cite{progressive-pcgan} and PDGN \cite{pdgn} progressively generate the point cloud from a low resolution to a high resolution. Inspired by StyleGAN \cite{stylegan}, MRGAN \cite{mrgan} and SP-GAN \cite{spgan} use AdaiN \cite{adain} for part-aware generation and shape editing. While SP-GAN manually chooses parts for editing, MRGAN learns to split shapes into parts in an unsupervised manner. Because the voxel requires large memory footprints and the point cloud lacks topological structures, researchers turn to deep implicit functions \cite{im-gan, deepsdf} to represent complex shapes with little memory cost. IM-GAN \cite{im-gan} designs a l-GAN to generate the occupancy values of the shapes. In IM-GAN, the GAN generates the shape code vectors which are further decoded to the occupancy values by a pre-trained decoder. Implicit-Grid \cite{grid-im-gan} proposes a grid-based implicit function for shape generation, which can capture more local details and enables the spatial control of the generation process. ShapeGAN \cite{shapegan} proposes an SDF-based GAN and studies how to combine the signed distance function (SDF) with the voxel-based or the point-based discriminator. GET3D \cite{get3D} generates meshes with textures and is trained with losses defined on rendered 2D images. SDF-styleGAN \cite{sdfstylegan} introduces the advanced StyleGAN2 \cite{stylegan2} to 3D shape generation, and proposes both global and local discriminators for the generated SDF values.

\noindent\textbf{Other generative methods.}
PQ-NET \cite{pq-net} proposes a sequence-to-sequence autoencoder to generate 3D shapes via sequential part assembly. PolyGen \cite{polygen} predicts the vertices and faces of the meshes sequentially by using a Transformer-based framework. GCA \cite{gca} introduces a new generative cellular automata to generate 3D shapes progressively. In addition, the flow-based generative model \cite{pointflow, dpf, softflow}, and the diffusion model \cite{luo2021diffusion, wu2023sketch_iccv} are also applied to the point cloud. 3DILG \cite{3Dilg} takes the transformer as the backbone and applies irregular grids to the implicit function, which allows shape codes to have arbitrary positions instead of pre-defined ones.

\begin{table}[]
	\begin{center}
		\resizebox{\linewidth}{!}
		{
			\begin{tabular}{lccc}
				\toprule
				Method & Representation & \makecell[c]{Part-level \\ Interpolation} & \makecell[c]{Semantic \\ Editing} \\ \midrule
				3D-GAN [2017] & voxel & \textcolor{red}{\ding{56}} & \textcolor{red}{\ding{56}} \\
				G2L-GAN [2018] & voxel & \textcolor{red}{\ding{56}}  & \textcolor{red}{\ding{56}} \\
				PAGENet [2020]  & voxel & \textcolor{green}{\ding{52}} & \textcolor{red}{\ding{56}} \\
				GCA [2021] & voxel & \textcolor{red}{\ding{56}} & \textcolor{red}{\ding{56}} \\
				PC-GAN [2018] & point cloud & \textcolor{red}{\ding{56}} & \textcolor{red}{\ding{56}} \\
				Tree-GAN [2019] & point cloud & \textcolor{red}{\ding{56}} & \textcolor{red}{\ding{56}} \\
				PointFlow [2019] & point cloud & \textcolor{red}{\ding{56}} & \textcolor{red}{\ding{56}} \\
				DPF-Net [2020] & point cloud & \textcolor{red}{\ding{56}} & \textcolor{red}{\ding{56}} \\
				MRGAN [2021] & point cloud & \textcolor{green}{\ding{52}} & \textcolor{red}{\ding{56}} \\
				SP-GAN [2021] & point cloud & \textcolor{green}{\ding{52}} & \textcolor{red}{\ding{56}} \\
				IM-GAN [2019] & implicit function & \textcolor{red}{\ding{56}} & \textcolor{red}{\ding{56}} \\
				ShapeGAN [2020] & implicit function & \textcolor{red}{\ding{56}} & \textcolor{red}{\ding{56}} \\
				Implicit-Grid [2021] & implicit function & \textcolor{red}{\ding{56}} & \textcolor{red}{\ding{56}} \\
				SDF-styleGAN [2022] & implicit function & \textcolor{red}{\ding{56}} & \textcolor{red}{\ding{56}} \\
				\hline
				\makecell[l]{3D Semantic \\ Subspace Traverser} & implicit function & \textcolor{green}{\ding{52}} & \textcolor{green}{\ding{52}} \\
				\bottomrule
			\end{tabular}
		}
	\end{center}
	\caption{Typical 3D generative models and their applications.}
	\label{tab:compare}
\end{table}

Our method combines a shape code grid~\cite{voxel-based-3D-detection-and-reconstruction-of-multiple-objects-from-a-single-image} with a novel l-GAN which consists of a variational autoencoder (VAE) and a GAN. Compared to IM-GAN and Implicit-Grid which also use the l-GAN, our continuous shape code grid naturally solves the discontinuity problem in the generated shapes by the trilinear interpolation. Different from the discrete grids in Implicit-Grid, ShapeFormer \cite{shapeformer}, and AutoSDF \cite{autosdf}, our continuous one is better at representing shapes with small grid resolution. Since the shape code grids produced by the VAE follow the Gaussian distribution, our GAN can easily learn to generate the shape code grids with complex structures. For shape editing, thanks to the spacial prior of the shape code grid and the local linear subspace model, we can do part-level interpolation and semantic editing. In \cref{tab:compare}, we summarize typical 3D generative models and their applications.

\subsection{Shape Editing}
% 这里参考下张浩老师的关于语义的文章。
The 3D part-aware generative methods, such as PQ-NET \cite{pq-net}, PAGENet \cite{pagenet}, MRGAN \cite{mrgan}, and SP-GAN \cite{spgan}, can do shape editing via part-level shape interpolation. PQ-NET and PAGENet learn the part segmentations from the part labels. MRGAN learns to disentangle the shape parts without any part-level supervision. In SP-GAN, a sphere prior provides the part information. There are some 2D-to-3D shape editing methods, such as Sketch2Mesh \cite{sketch2mesh} and MM-VADs \cite{mmvad}. Sketch2Mesh uses an encoder-decoder network to reconstruct shapes from sketches and refines 3D shapes through differentiable rendering. MM-VADs proposes a general multimodel generative model to edit and reconstruct 3D shapes from 2D sketches. Both of them can manipulate a 3D shape by editing its corresponding 2D sketch. DIF-NET \cite{dif-net} proposes a deformed implicit field representation, which builds the dense correspondences among shapes via a template. Thanks to the dense correspondences, DIF-NET can edit a shape by moving a 3D point to a new position. Similarly, DualSDF \cite{dualsdf} proposes a new representation with two levels of granularity, i.e., the SDF and the primitive. DualSDF can edit the SDF of the shape by moving the primitives. SPAGHETTI \cite{SPAGHETTI} can directly edit neural implicit shapes and enable part-level control by applying transforms, blending, and inserting fragments of different shapes to local regions of the generated object. ShapeCrafter \cite{shapecrafter} utilizes text descriptions to recursively edit shapes.

There are some non-generative part-aware 3D shape editing methods, such as Wei et al.\cite{wei2020learning} and DeepMetaHandles\cite{deepmetahandles}. Wei et al.\cite{wei2020learning} relies on pre-defined semantic attributes, limiting editing possibilities. DeepMetaHandles\cite{deepmetahandles} learns to move a set of control points, lacking the ability to add or remove parts.

For shape editing, our method does not need any part-level supervision or dense correspondences. We utilize the spacial prior of the shape code grid and the proposed local linear subspace model to discover the semantic dimensions of the local regions. By traversing those semantic dimensions, we can edit the semantic attributes of the shapes, such as adding or removing parts, changing the type of legs, and so on.

\subsection{Interpretability Learning for GANs}
GANSpace \cite{ganspace} uses principal component analysis (PCA) in the feature space to identify the important latent directions, which represent the interpretable variations. SeFa \cite{sefa} tries to find the semantic directions by decomposing the weights of the pre-trained generator. EigenGAN \cite{eigengan} embeds the linear subspace models to the intermediate feature maps of the generator to discover the semantic dimensions. By traversing those dimensions, EigenGAN can edit the semantic attributes of the outputs.

The methods mentioned above can only learn to control the global semantic attributes. In contrast, we propose the local linear subspace model to learn both the local and global semantic attributes of a feature map. We use differentiable embedding to embed the local linear subspace model to a local region of the feature map. The size of the region we embed determines whether the subspace model learns local or global attributes.

\section{Method}

\begin{figure*}[t]
	\centering
	\includegraphics[width=\linewidth]{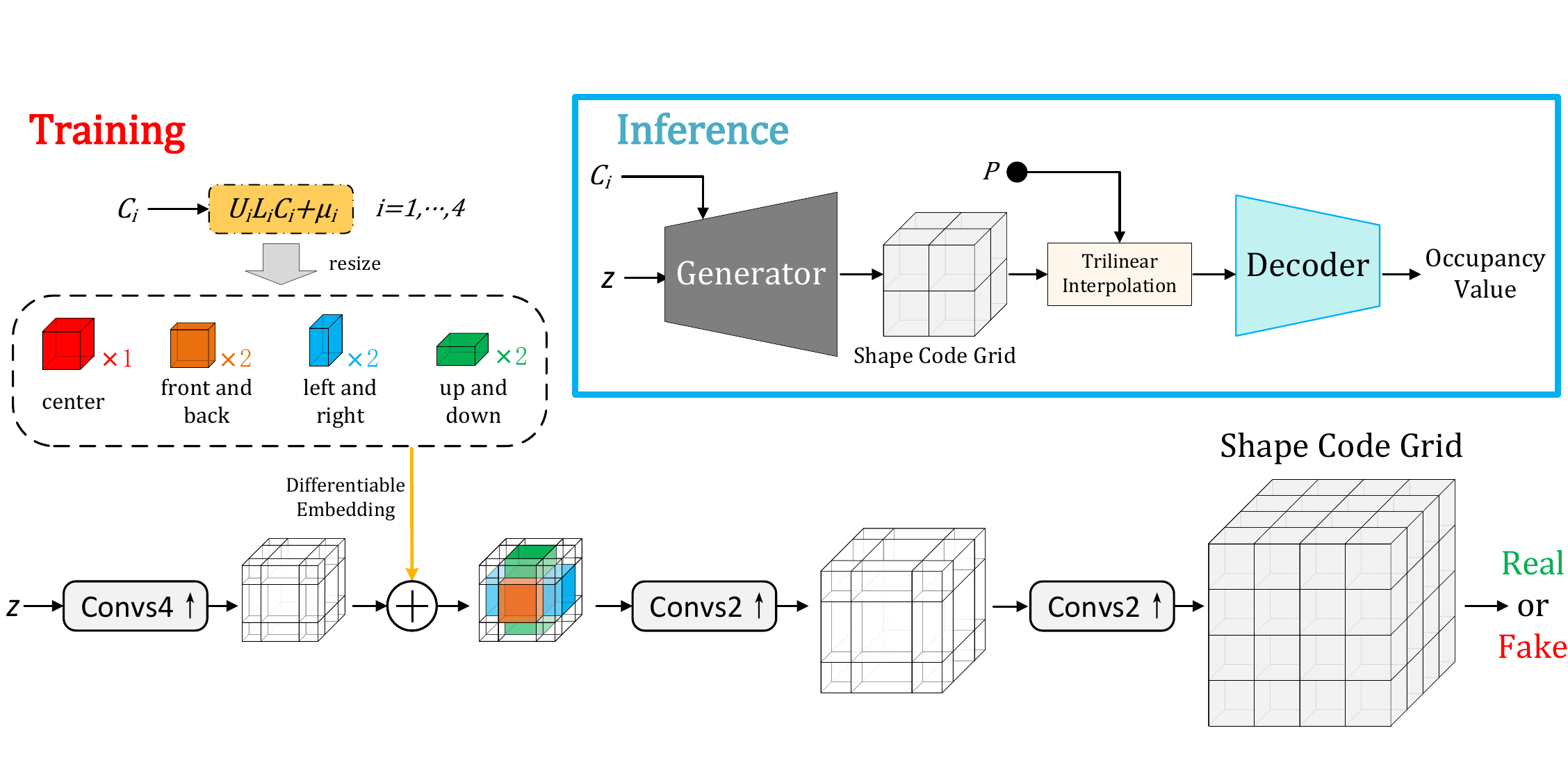}
	\caption{The overview of our 3D Semantic Subspace Traverser. $z$ and $C_i$ are random vectors. Convs$X$ represents the transposed convolutional block with stride $X$. To learn the semantic information, we embed four local linear subspace models (the red cube, the blue cubes, the green cubes, and the orange cubes) into the feature map by differentiable embedding. It is important to note that two orange cubes (as well as blue and green ones) are combined into a single subspace, while they are embedded in distinct regions. It is viable by using transform matrices containing two identity matrices. While the resolution of the feature map is $r^3$, the edge length of the red cube is $\frac{1}{2}r$. We embed the red cube into the center of the feature map, and other cubes are embedded into the positions beside the red cube. We present a brief inference process in the top right corner to show how we generate and edit shapes.}
	\label{fig:generator}
\end{figure*}

\subsection{Overview}
We design 3D Semantic Subspace Traverser, a novel semantic 3D generative model for 3D shape generation and editing, as illustrated in \cref{fig:generator}. In the following sections, we first introduce our local linear subspace model and how to embed it into the local regions of the feature map by differentiable embedding in \cref{sec:subspace}. Second, we present the details of our 3D Semantic Subspace Traverser in \cref{sec:generator}. Then, we talk about how we encode shapes into the latent shape code grids in \cref{sec:vae}. Lastly, we present the loss function and implementation details in \cref{sec:loss} and \cref{sec:impl_detail} respectively.

\subsection{Local Linear Subspace Model}
\label{sec:subspace}

Different from the linear subspace model in EigenGAN \cite{eigengan}, our local linear subspace models can be embedded into local regions to learn semantics. In the forward propagation, we first take a random vector $C=[c_1,\cdots,c_n]^T$ as the coordinates to sample a point $\phi \in R^{c \times d\times h\times w}$ from the local linear subspace:
\begin{equation}
  \phi = ULC + \mu = \sum_{i=1}^{n}u_il_ic_i + \mu
  \label{eq:point}
\end{equation}
Here $U=[u_1,\cdots,u_n], u_i \in R^{c \times d\times h\times w}$ are the orthonormal bases. $n$ is the number of bases. $L=diag(l_1,\cdots,l_n)$ is a diagonal matrix in which $l_i$ is the importance of the basis vector $u_i$.  $\mu$ is the origin of the subspace.

\begin{figure}[t]
	\centering
	\includegraphics[width=\linewidth]{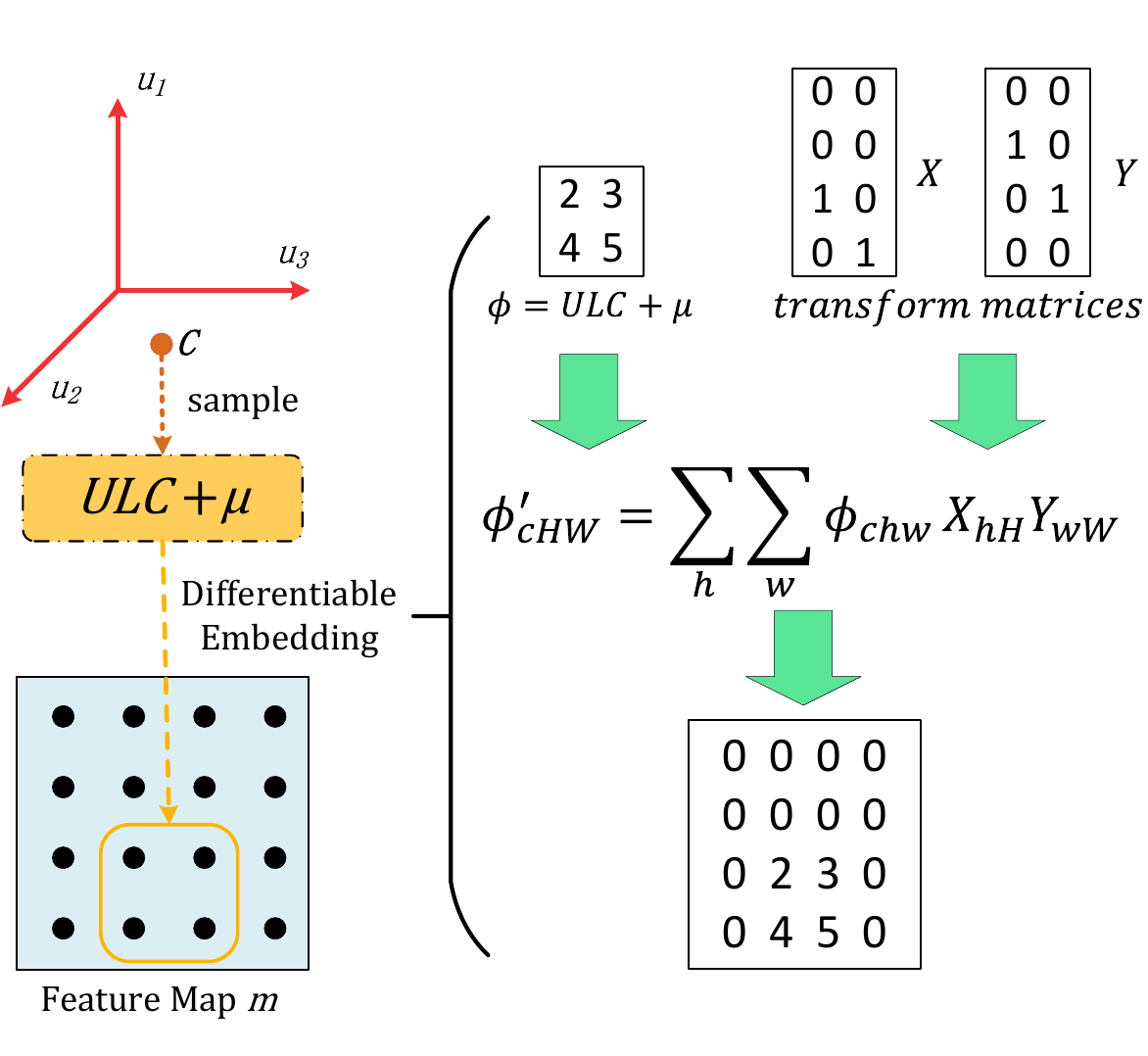}
	
	\caption{2D Example of the embedding process. We first sample a point $\phi$ from the subspace. Then to make the embedding process differentiable, we use transform matrices instead of indexing to embed the point $\phi$ into the specified local region of the feature map $m$.}
	\label{fig:subspace}
\end{figure}

Then, we use differentiable embedding to embed this point $\phi$ into the local region of the feature map $m \in R^{c \times D\times H\times W}$:
\begin{equation}
  \hat{m} = m + f(\phi, X, Y, Z) \\
  \label{eq:embedding}
\end{equation}
\begin{equation}
  f(\phi, X, Y, Z)=\sum_{d} \sum_{h} \sum_{w} \phi_{cdhw} X_{dD} Y_{hH} Z_{wW} \\
  \label{eq:einsum}
\end{equation}
Since the sizes of $m$ and $\phi$ are not the same, embedding by indexing is not differentiable. To solve this problem, we use a padding function $f$ to expand the size of $\phi$ for differentiable embedding. As shown in  \cref{eq:einsum}, we express the tensor calculations by Einstein notation. $X \in R^{d \times D}$, $Y \in R^{h \times H}$, and $Z \in R^{w \times W}$ are transform matrices that consist of a square identity matrix and zeroes, like
$
\left[\begin{array}{ccc}
	0 & I & 0
\end{array}\right]^T
$.
Those transform matrices expand different dimensions of the $\phi$ respectively and control the position of the $\phi$ after padding.
In other words, they decide which region of the feature map $m$ we embed the $\phi$ into. By this means, we replace the indexing with the transform matrices, which makes the embedding differentiable. \cref{fig:subspace} shows a 2D example to illustrate the differentiable embedding. 
All the parameters of the local linear subspace model $S=(U,L,\mu)$ are learnable and are trained with the holistic 3D Semantic Subspace Traverser network. 

After training, the local linear subspace model captures the semantic attributes of the local region in the feature map $m$. We can semantically manipulate the feature map by changing the coordinates $C$ of the sample point in the subspace.

\subsection{3D Semantic Subspace Traverser}
\label{sec:generator}

\cref{fig:generator} shows the overview of our 3D Semantic Subspace Traverser. We embed four local linear subspace models into the feature map $m$ between the generator's layers to control the semantic attributes as:
\begin{equation}
  \hat{m} = m + \sum_{i=1}^{4}g(f(\phi_i, X_i, Y_i, Z_i)) \\
  \label{eq:gan_subspace}
\end{equation}
Here $g$ is an identity function or a simple convolution. The embedding layout is illustrated in \cref{fig:generator}. We design a centrosymmetric layout according to the symmetry of training data. In this layout, the center subspace makes global changes to shapes, and other subspaces edit local shape parts. Such an embedding layout, in our opinion, is generally effective and can manipulate both local and global regions of shapes. 

Our generator takes a random noise $z \sim N(0, 1)$ and a set of random vectors $\left\{ C_i | C_i \sim N(0, 1) , i=1, \dots, 4 \right\}$ as input. $C_i$ is the coordinates for the $i^{th}$ local linear subspace to sample a point.  $G(z, C_i)$ is the generated shape code grid.

We design a 3D PatchGAN discriminator $D$ to focus on the patches of the shape code grid. The discriminator follows the basic idea of the 2D PatchGAN discriminator from Pix2Pix \cite{pix2pix}. It classifies some overlapping $N \times N \times N$ patches in the shape code grid as real or fake. By this means, we penalize the shape code grid at the scale of patches to improve the diversity of the generated shapes. More network details of the GAN are provided in the supplementary material.

At the inference time, the generated shape code grid is decoded into occupancy values by a pretrained implicit decoder, which we will talk about in \cref{sec:vae}. And we can extract the mesh from the occupancy values by marching cubes algorithm \cite{marchingcubes}.

\subsection{Shape Encoding}
\label{sec:vae}

\begin{figure}[t]
	\centering
	\includegraphics[width=\linewidth]{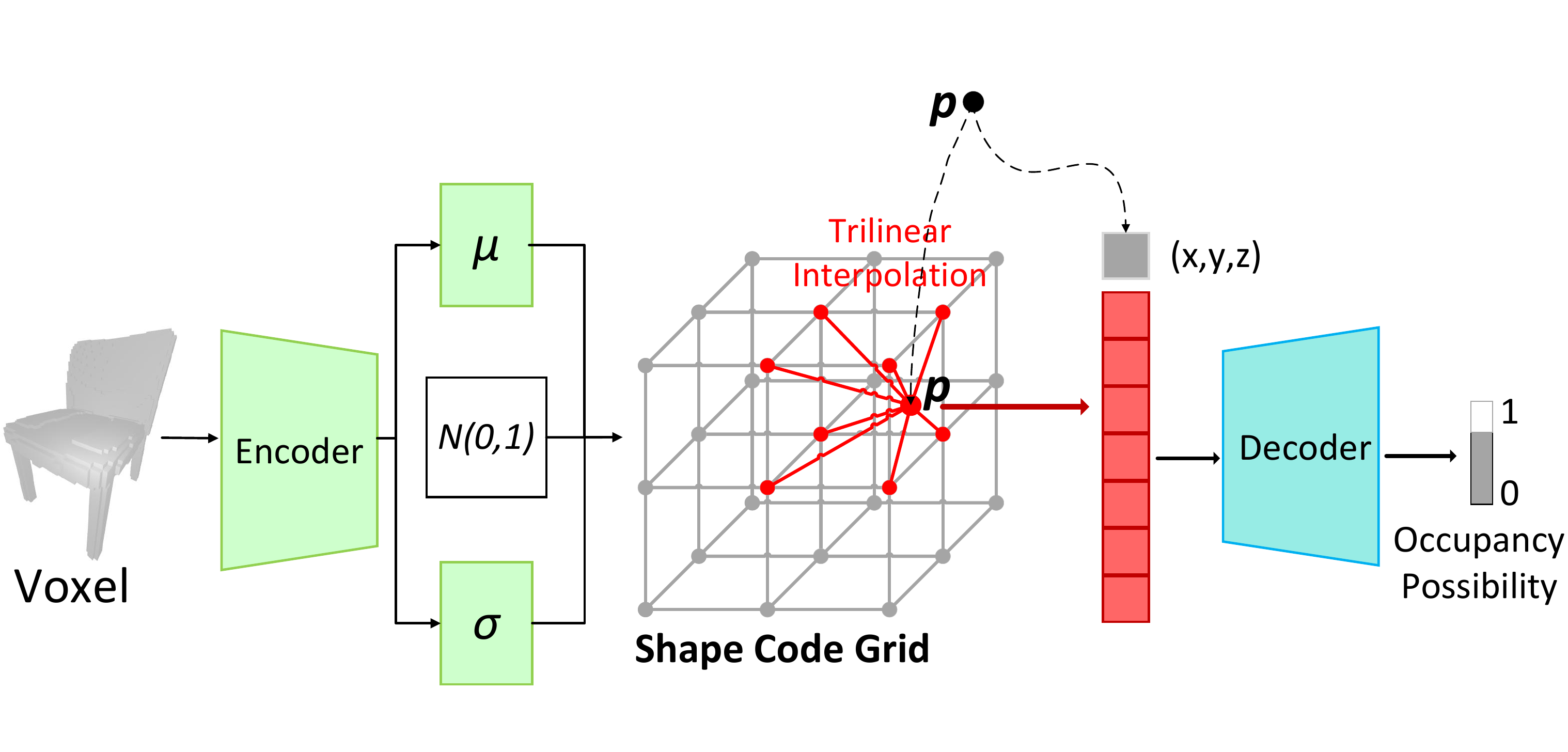}
	\caption{The VAE for shape encoding. In this figure, we take a $3^3$ shape code grid as an example. Seven red squares are the shape features of the point $\textbf{p}$ and its six surrounding points.}
	\label{fig:vae}
\end{figure}

To train our 3D Semantic Subspace Traverser, we take a VAE $V$ to encode 3D shapes into the latent shape code grids. The holistic framework of our VAE is shown in \cref{fig:vae}. We train the VAE in a voxel-superresolution manner. The encoder first encodes $r^3$ voxels into the mean $\mu$ and the standard deviation $\sigma$ of a distribution. Then we sample a shape code grid $s \in R^{c \times g  \times g  \times g}$ from this distribution by reparameterization. $c$ and $g$ are the channel and resolution of the shape code grid respectively.

The decoder takes a point $\textit{\textbf{p}}$ and a shape code grid $s$ as the inputs to predict the occupancy values of the point $\textit{\textbf{p}}$.
First, we extract the shape features $F(\textit{\textbf{p}})$ from the grid at location $\textit{\textbf{p}}$ by using trilinear interpolation as shown in \cref{fig:vae}. Then, in order to get the local features around the location $\textit{\textbf{p}}$, we also extract the shape features $F(\textit{\textbf{p}}_{i})$ of six surrounding points $\textit{\textbf{p}}_{i}$. Those surrounding points are in a distance $d$ from $\textit{\textbf{p}}$ along three Cartesian axes. We concatenate the shape features $F(\textit{\textbf{p}})$, the local features $F(\textit{\textbf{p}}_{i})$, and the coordinates of $\textit{\textbf{p}}$ together, and feed them to the implicit decoder to get the occupancy value of $\textit{\textbf{p}}$. 

Once trained, the shape code grid $s$ produced by the encoder is used to train our 3D Semantic Subspace Traverser. More network details of the VAE are provided in the supplementary material.

\subsection{Model Learning}
\label{sec:loss}

The loss functions of our 3D Semantic Subspace Traverser are the non-saturating loss $\mathcal{L}_{GAN}$ \cite{gan}, the $R_1$ regularization $\mathcal{R}_{1}$ \cite{r1}, and the subspace regularization loss $\mathcal{L}_{subspace}$ \cite{eigengan}. To be specific, the loss of the generator is
\begin{equation}
	\mathcal{L}_{G}=\mathbb{E}[\zeta(-D(G(z, C_i)))] + \alpha \mathcal{L}_{subspace} \\
\end{equation}
Here $G(\cdot)$ and $D(\cdot)$ are outputs of the generator and discriminator respectively. $\mathcal{L}_{subspace}=\parallel U^TU - I \parallel_{F}^2$. $\alpha$ is the weight of $\mathcal{L}_{subspace}$ and $\alpha = 1$. $\zeta(x)=\log(1+\exp(x))$.

The loss of the discriminator is
\begin{equation}
	\begin{aligned}
		\mathcal{L}_{D} &= \mathbb{E}[\zeta(D(G(z, C_i)))] + \mathbb{E}[\zeta(D(s))]  \\
	     &+ \beta \mathcal{R}_{1}(s,\Theta_D)
	\end{aligned}
\end{equation}
Here $s$ is the shape code grid produced by the encoder of VAE. $\Theta_D$ are the parameters of the discriminator $D$. $\beta$ is the weight of $ \mathcal{R}_{1}$ and $\beta = 1$.

We train our VAE by the binary cross-entropy loss and the KL loss:
\begin{equation}
	\begin{aligned}
		\mathcal{L}_{VAE} &=\mathcal{L}_{BCE}(V(a, \textit{\textbf{p}}), o) \\
		 &+ KL(N(\mu, \sigma)||N(0, 1))
	\end{aligned}
\end{equation}
Here $\mathcal{L}_{BCE}$ is the binary cross entropy loss. $a$ is the input voxel. $V(\cdot)$ is the predicted occupancy values of VAE. $o$ are the ground truth occupancy values. $KL$ is the KL loss.$\mu$ and $\sigma$ are the mean $\mu$ and the standard deviation produced by the encoder of VAE.

\subsection{Implementation Details}
\label{sec:impl_detail}
In 3D Semantic Subspace Traverser, we set $n=6$, which is the number of dimensions of all subspace models. In experience, when $n \leq 3$, the subspace models have difficulty in learning semantics. When $n > 3$, the subspace model is able to capture semantics. $n=6$ is chosen as a trade-off between the learned semantics and the number of parameters.
In the VAE, we set $r=64$, $c=8$, $g=16$, and $d=0.0536$.

\section{Experiments}
\label{sec:expeiments}

\subsection{Dataset and Details} 
We use the ShapeNet Core (v1) dataset \cite{shapenet2015} to train our network. For comparison, we adopt the data split of  \cite{sdfstylegan} \cite{hsp}: 70\% of the data for training, 20\% of the data for testing, and the rest for validating which is not used in our experiments. For data preprocessing, we follow the steps in  \cite{if-net}. To be specific, we first convert the meshes to watertight ones. Second, we normalize those meshes to make the lengths of their largest bounding box edges equal to one. Third, we sample some points and compute their occupancy values for each mesh. Finally, we voxelize meshes to produce $64^3$ voxels. Our VAE is trained with five categories of the ShapeNet Core (v1) dataset, including the chair, airplane, table, car, and rifle. Different from the VAE, we train one specific GAN for each specific category. In other words, our GAN is category-specific.

In the following subsections, we first present the shape editing results in \cref{sec:shape edit}. Then, we show the quantitative evaluation of our method and other 3D generative methods in \cref{sec:comparision}. Finally, we present the voxel super-resolution experiment in \cref{sec:recon}.

\subsection{Semantic Attributes for Shape Editing}  % 标题最好用名词
\label{sec:shape edit}

\begin{figure}[t]
	\centering
	\includegraphics[width=\linewidth]{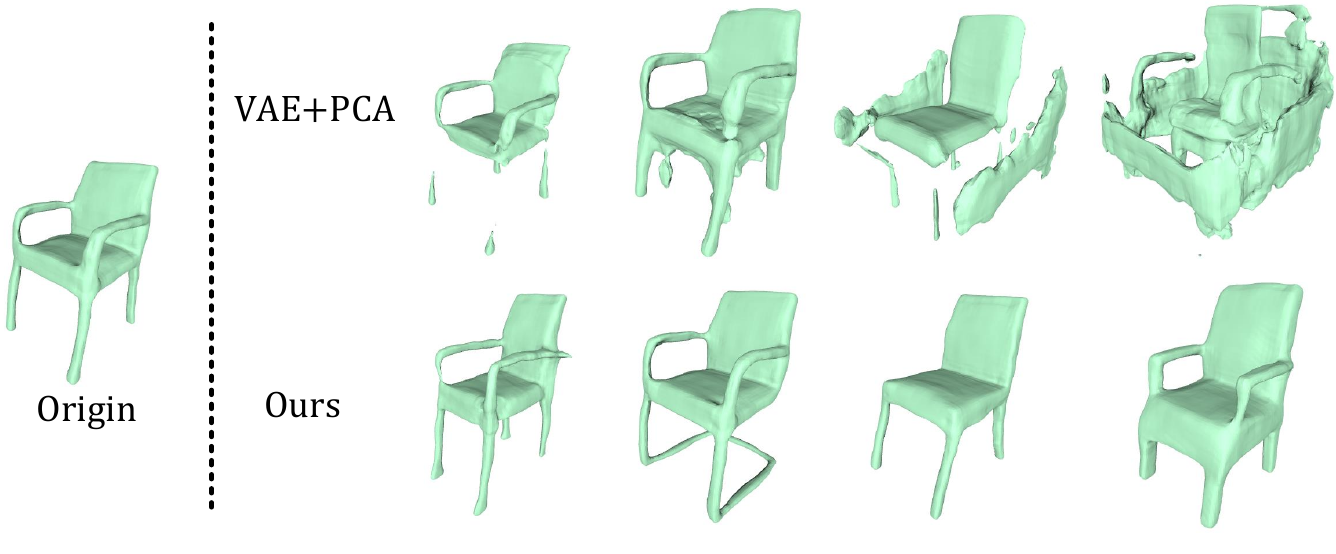}
	\caption{Editing Comparision with VAE+PCA. We use our method and VAE+PCA to edit the origin shape and produce four results.}
	\label{fig:pca}
\end{figure}

\begin{figure*}[t]
	\centering
	\includegraphics[width=0.975\linewidth]{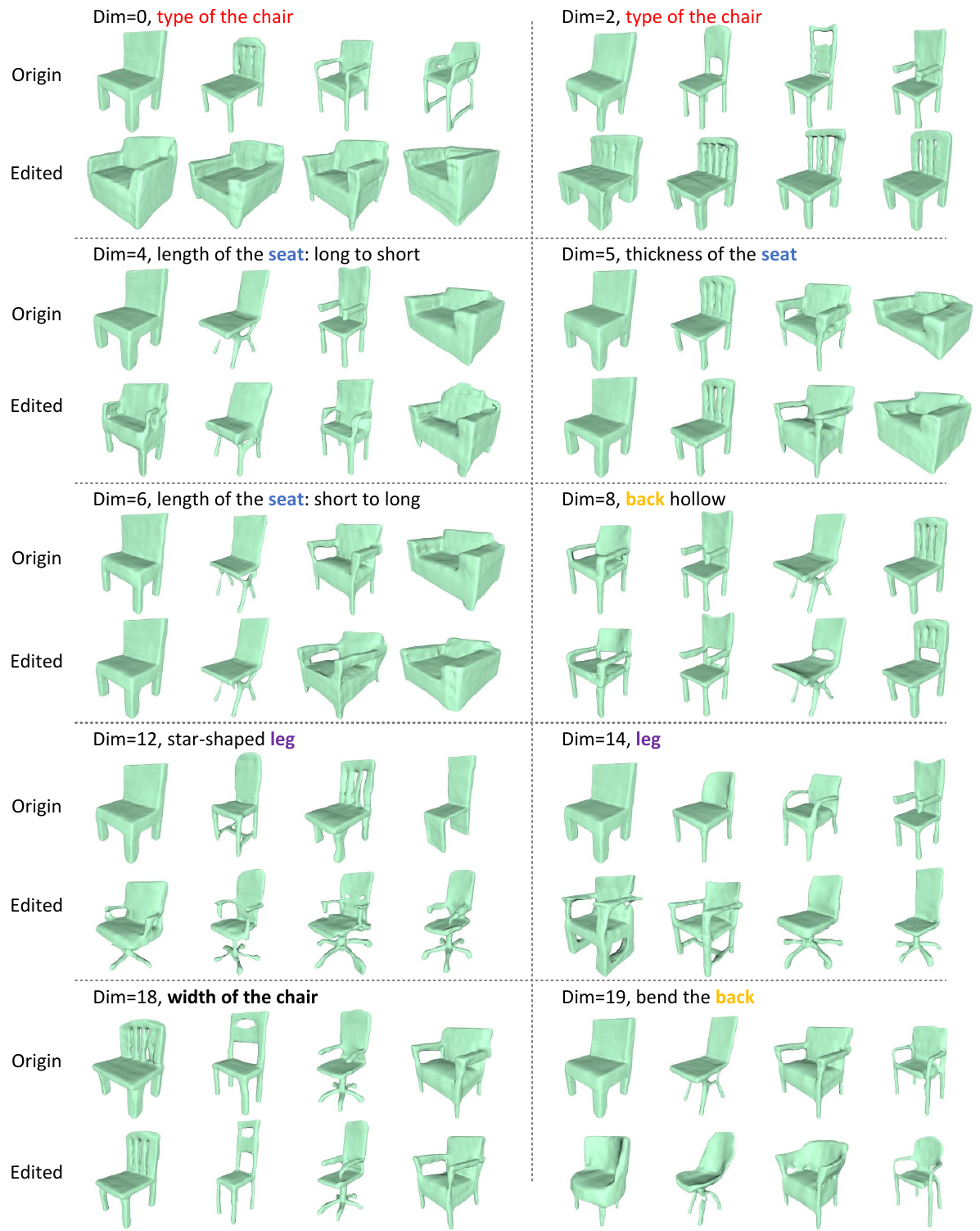}
	\caption{
		The learned semantic attributes of the local linear subspaces in the chair category. ``Dim=i” denotes the $\rm i^{th}$ dimension of all local linear subspace models.}
	\label{fig:edit1}
\end{figure*}

\begin{figure*}[t]
	\centering
	\includegraphics[width=\linewidth]{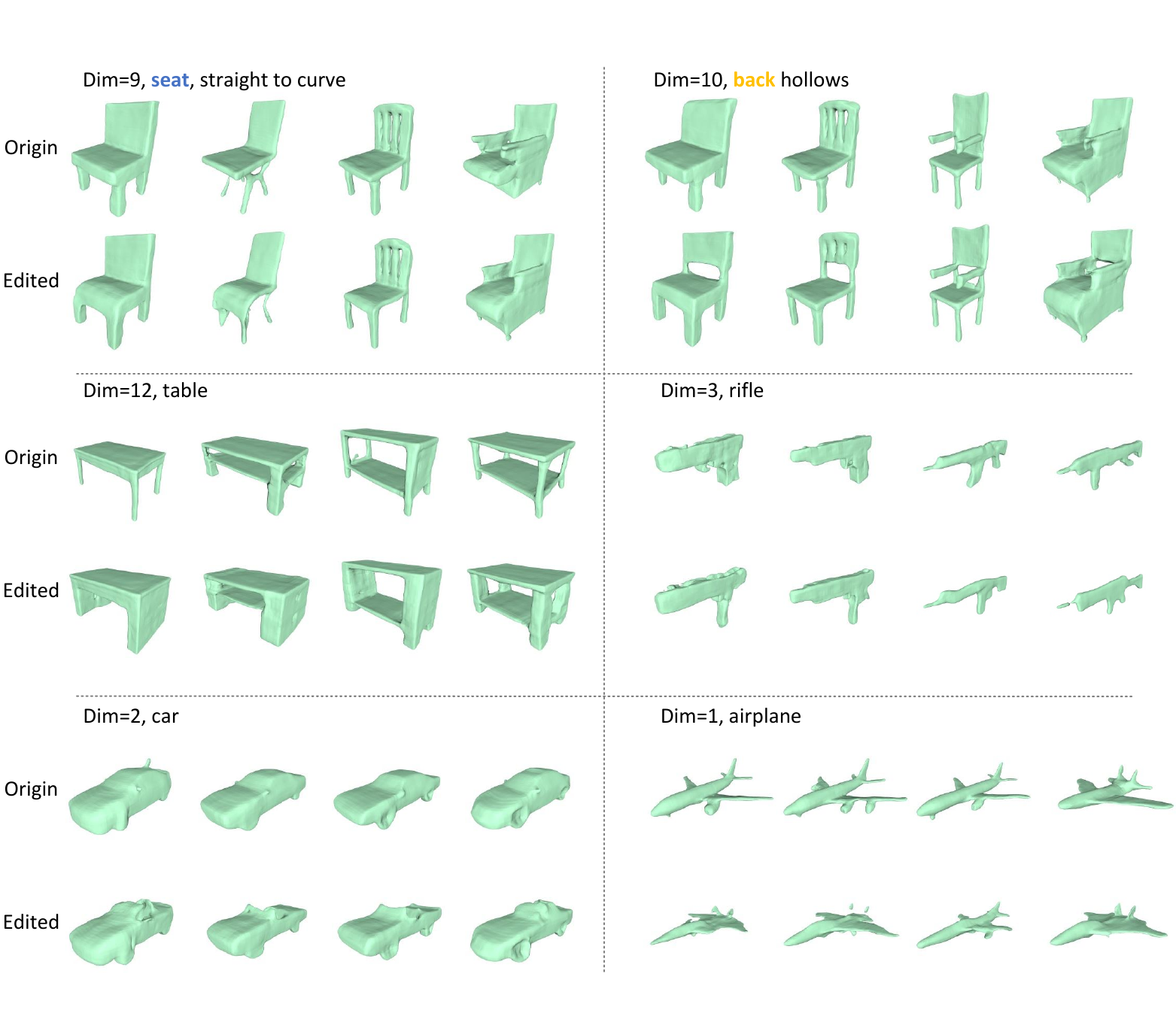}
	\caption{
		The learned semantic attributes of the local linear subspaces in various categories. ``Dim=i” denotes the $\rm i^{th}$ dimension of all local linear subspace models.}
	\label{fig:edit2}
\end{figure*}

We show various shape edit results in \cref{fig:edit1} and \cref{fig:edit2}, where ``Dim=i” is the $\rm i^{th}$ dimension of all local linear subspace models. In five categories, different kinds of semantic attributes are learned by different dimensions. As shown in \cref{fig:edit1} and \cref{fig:edit2}, moving along those dimensions, chairs' semantic attributes of the generated shapes change. We can edit both local attributes, e.g., ``seat" in Dim = 4, 5, 6, 9, ``back" in Dim = 8, 10, 19, and ``leg" in Dim = 12, 14; and global attributes, e.g., ``type" in Dim = 0, 2, and ``width" in Dim = 18. Specifically, when we traverse along the $\rm 4^{th}$ and $\rm 6^{th}$ dimensions, the same attributes are edited but with opposite results. The $\rm 12^{th}$ and $\rm 14^{th}$ dimensions show similar results. The $\rm 12^{th}$ dimension translates all chairs to the ones with star-shaped legs, while the $\rm 14^{th}$ dimension changes all legs to two types. Our method maintains the right semantic structures without inconsistency after editing shape parts. Since our training is unsupervised, there may be some levels of semantic entanglement. Despite this, each dimension typically maintains a dominant semantic attribute. More importantly, the resulting edits are reasonably coherent and surpass the outcomes produced by the VAE+PCA method in \cref{fig:pca}. We will discuss this comparison in more detail later.

\cref{fig:edit2} shows semantic attributes learned from different categories. For example, we transfer airliners to fighter planes by traversing the $\rm 1^{th}$ dimension. Moreover, it is noticeable that certain types of data, such as chairs, exhibit more pronounced semantic structure changes compared to rifles or cars. It is because our method leverages subspace models, which excel at capturing these larger structural variations, leading to stronger visual results with the categories like the chair. 

\noindent\textbf{Comparision with VAE+PCA.} We provide a meaningful baseline for comparison, which is to apply principal components analysis (PCA) to the VAE encoder output space to discover semantic dimensions. In \cref{fig:pca}, we show some editing results of our method and VAE+PCA. As illustrated, VAE+PCA typically generates unreasonable float artifacts, whereas our results preserve the proper semantic structures. In conclusion, our approach captures semantic information more effectively than PCA.

\subsection{Quantitative Evaluation of Shape Generation}
\label{sec:comparision}
To evaluate our shape generation quality, we compare our method to other representative methods that use implicit functions for 3D generation: IM-GAN \cite{im-gan}, ShapeGAN \cite{shapegan}, Implicit-Grid \cite{grid-im-gan}, and SDF-StyleGAN \cite{sdfstylegan}. Among these methods, IM-GAN, Implicit-Grid, and our method use occupancy values to represent 3D shapes. ShapeGAN and SDF-StyleGAN use SDF values to represent 3D shapes. 

For a fair comparison, we take ECD \cite{grid-im-gan} and Shading-image-based FID (S-FID) \cite{sdfstylegan} as our evaluation metrics, and also report COV \cite{pc-gan} and MMD \cite{pc-gan} values as the reference. Following the experiments setting in \cite{sdfstylegan}, ECD, MMD, and COV are all based on the light-field-distance (LFD) \cite{lfd, sdfstylegan} to measure the distance between two samples. LFD is based on silhouette images and measures the structure similarity between two shapes. When computing COV, MMD, and ECD, we take the testing dataset as the reference to compare with our generated shapes, which is five times the number of the reference. When computing S-FID, we use the training dataset as the reference, and the number of generated shapes is the same as the reference. We use the clean-fid algorithm \cite{clean-fid} to compute the FID.

We show the quantitative results in \cref{tab:metrics}. COV is listed for reference only as mentioned before. In all categories, our method performs best in MMD. And for three classes with complex topological structures, i.e., chair, airplane, and table, we also achieve the best results in ECD. It shows that our method is good at generating shapes with complex topological structures. But in S-FID, we are not the best due to the unsmooth surfaces.

\begin{table}[]
\begin{center}
\resizebox{\linewidth}{!}
{
\begin{tabular}{c|c|cccc}
\toprule
\textbf{Data}             & \textbf{Method} & \textbf{COV(\%)↑} & \textbf{MMD↓} & \textbf{ECD↓} & \textbf{S-FID↓} \\ \midrule
\multirow{5}{*}{\rotatebox{90}{Chair}} & IMGAN         & 72.57          & 3326          & 1998         & 63.42          \\
                       & Implicit-Grid & \textbf{82.23} & 3447          & 1231         & 119.5          \\
                       & ShapeGAN      & 65.19          & 3726          & 4171         & 126.7          \\
                       & SDF-StyleGAN  & 75.07          & 3465          & 1394         & \textbf{36.48} \\
                       & Ours          & 78.61          & \textbf{3160} & \textbf{810} & 99.64          \\ \midrule
\multirow{5}{*}{\rotatebox{90}{Airplane}} & IMGAN           & 76.89             & 4557          & 2222          & 74.57         \\
                       & Implicit-Grid & \textbf{81.71}          & 5504          & 4254         & 145.4          \\
                       & ShapeGAN      & 60.94          & 5306          & 6769         & 162.4          \\
                       & SDF-StyleGAN  & 74.17          & 4989          & 3438         & \textbf{65.77}          \\
                       & Ours          & 73.42          & \textbf{3879}          & \textbf{1933}         & 133.4               \\ \midrule
\multirow{5}{*}{\rotatebox{90}{Car}}   & IMGAN         & 54.13          & 2543          & 12675        & 141.2          \\
                       & Implicit-Grid & \textbf{75.13}          & 2549          & 8670         & 209.3          \\
                       & ShapeGAN      & 57.40          & 2625          & 14400        & 225.2          \\
                       & SDF-StyleGAN  & 73.60          & 2517          & \textbf{6653} & \textbf{97.99}          \\
                       & Ours          & 64.67          & \textbf{1509} & 10355             & 219.5               \\ \midrule
\multirow{5}{*}{\rotatebox{90}{Table}} & IMGAN         & 83.43          & 3012          & 907          & 51.70          \\
                       & Implicit-Grid & \textbf{85.66} & 3082          & 1089         & 87.69          \\
                       & ShapeGAN      & 76.26          & 3236          & 1913         & 103.1          \\
                       & SDF-StyleGAN  & 69.80          & 3119          & 1729         & \textbf{39.03} \\
                       & Ours          & 79.78          & \textbf{2698} & \textbf{539} & 99.95          \\ \midrule
\multirow{5}{*}{\rotatebox{90}{Rifle}} 
& IMGAN         & 71.16          & 5834          & 701          & 103.3          \\
                       & Implicit-Grid & 77.89          & 5921          & \textbf{357} & 125.4          \\
                       & ShapeGAN      & 46.74          & 6450          & 3115         & 182.3          \\
                       & SDF-StyleGAN  & \textbf{80.63} & 6091          & 510          & \textbf{64.86}          \\
                       & Ours          & 62.73          & \textbf{4251} & 1207         & 154.4               \\ \bottomrule
\end{tabular}
}
\end{center}
\caption{The quantitative evaluation of different 3D generation methods.}
\label{tab:metrics}
\end{table}

\subsection{Voxel Super-Resolution}
\label{sec:recon}
This section evaluates our shape compression ability by the voxel super-resolution experiment. We compare our VAE with IM-NET \cite{im-gan} and IF-NET \cite{if-net} (see in \cref{tab:recon}). For a fair comparison, we use five classes in the data split of IF-NET, which are chairs, airplanes, tables, cars, and rifles. We train all models for 200 epochs. All methods take $32^{3}$ voxels as inputs, and the output meshes are sampled at $128^3$. To measure the quality, we employ three recognized measures \cite{if-net}: volumetric intersection over union (IoU) evaluates how well the defined volumes overlap, Chamfer-L2 evaluates the accuracy and completeness of the shape surface, and normal consistency evaluates the accuracy and completeness of the shape normals. \cref{tab:recon} shows that our method achieves the second-best results, which is reasonable. In contrast to IM-NET, our method utilizes a grid to represent the shape code rather than a vector, which improves the quality of local regions' reconstruction. And IF-NET outperforms us in terms of metrics since it utilizes multi-scale features. Our method makes a balance between quality and network complexity. Additionally, our main contribution is in the GAN part. Based on \cref{tab:recon}, we believe that the current VAE is good enough to support our method.
% 数据，网络的输入分辨率、输出什么（Occupancy到obj文件），epoch多少，指标是什么，列举表格，简单说明（指标代表什么，我们的方法与他的方法的差别。如果我们的方法比IF-NET好，那么可能是epoch的原因，IF-NET的原设置epoch相当长，而我们这里只使用了200。）。

\begin{table}[t]
\begin{center}
    \begin{tabular}{c|c|c|c}
    \toprule
        Method & IoU $\uparrow$ & Chamfer-$L_{2}\times 10^2 \downarrow$ &  \begin{tabular}[c]{@{}c@{}}Normal-\\ Consis.\end{tabular} $ \uparrow $ \\ \midrule
        IM-NET & 0.754 & 0.160 & 0.877 \\ 
        IF-NET & \textbf{0.907} & \textbf{0.002} & \textbf{0.960} \\ 
        Ours & 0.804 & 0.107 & 0.925 \\ 
    \bottomrule
    \end{tabular}
\end{center}
\caption{Quantitative Evaluation of voxel super-resolution.}
\label{tab:recon}
\end{table}

\section{Conclusion and Discussion}
\noindent\textbf{Conclusion.} In this paper, we introduce a novel semantic 3D generative model, 3D Semantic Subspace Traverser, which can edit semantic attributes of 3D shapes. We apply a 3D Semantic Subspace Traverser to the latent shape code and use our local linear subspace models to capture the semantic information. Once trained, each dimension of subspaces corresponds to a specific semantic attribute, and we can edit shapes by traversing along those dimensions.

% Our new type of l-GAN is composed of a VAE and a GAN, in which the latent code of the VAE can be easily learned by the GAN. Thanks to the format of the shape code grid and our new type of l-GAN, we can generate smooth shapes with complex structures. What is more, the local linear subspace models, which we embed to the feature maps of the generator, can capture the semantic information of shapes and control their semantic attributes for shape editing. We show that our method can edit the local and global semantic attributes of shapes. Unlike previous methods based on part-level interpolation, we can edit shape parts while maintaining reasonable semantic structures.

% 标题可能之后要改一下，考虑不用Limitation
% 1. 我们只能进行无监督的训练，得到的编辑结果并不确定。
% 2. 嵌入方式，可以多多变化，根据任务。
% 3. 怎么自动选择有意义的dimension。可以用分割，也可以直接看L（当然，L的值只能看出哪几个值会对最终形状产生影响）
\noindent\textbf{Limitations $\&$ Discussion.} The limitation of our method is the handcrafted embedding layout. It would be our future work to design a mechanism that can automatically choose the embedding layout.

And for the current handcraft embedding layout, we provide a generally effective design in \cref{fig:generator} for users. It is also free to propose different embedding layouts for different tasks. There is some advice for designing the embedding layout. First, depending on where you wish to learn the semantics, you would do better to pick the suitable subspace size and embedding locations. Second, the number of dimensions in a subspace should be greater than three; otherwise, it is hard to learn the semantics.

Automatic dimension selection for editing is still an open issue in the literature. A possible solution is based on 3D segmentation: first, segment an original shape and its corresponding edited shape into semantic parts, and then compare corresponding parts to decide whether the semantic dimension is meaningful. Moreover, since our learned attributes are unpredictable, taking texts as guidance to lead dimensions to learn specific semantic attributes is a viable solution. These are also among our future work.

% In the future, we plan to design an adaptive region selection mechanism to replace the manual selection. Also, we would like to explore how to combine our local linear subspace model with other kinds of generative models. Style transfer of 3D shapes is also an interesting topic for shape editing.
\paragraph{Acknowledgement.}
The research was sponsored by the National Natural Science Foundation of China (No. 62176170).

%%%%%%%%% REFERENCES
{\small
\bibliographystyle{ieee_fullname}
\bibliography{egbib}
}

\end{document}